\documentclass{article}

\usepackage{nac_preprint}
\usepackage[utf8]{inputenc} 
\usepackage[T1]{fontenc}    
\usepackage{hyperref}       
\usepackage{url}            
\usepackage{booktabs}       
\usepackage{amsfonts}       
\usepackage{nicefrac}       
\usepackage{microtype}      

\usepackage{booktabs} 
\usepackage{subfigure}
\usepackage{amsmath}
\usepackage{amssymb}
\usepackage{cleveref}
\usepackage{mathtools}
\usepackage[toc,page]{appendix}
\usepackage{enumitem}
\usepackage{graphics}
\usepackage{graphicx}
\usepackage{xcolor}
\usepackage[export]{adjustbox}
\usepackage{algorithm}
\usepackage{algorithmicx}
\usepackage[noend]{algpseudocode}
\algnewcommand{\LineComment}[1]{\State \(//\) #1}
\algnewcommand{\RLineComment}[1]{\State \(\triangleright\) #1}
\usepackage{bm}
\usepackage{multirow}

\usepackage{wrapfig}

\usepackage{etoolbox}
\usepackage{tikz}
\usetikzlibrary{tikzmark}
\usetikzlibrary{calc}

\errorcontextlines\maxdimen

\newcommand{\ALGtikzmarkcolor}{black}
\newcommand{\ALGtikzmarkextraindent}{4pt}
\newcommand{\ALGtikzmarkverticaloffsetstart}{-.5ex}
\newcommand{\ALGtikzmarkverticaloffsetend}{-.5ex}
\makeatletter
\newcounter{ALG@tikzmark@tempcnta}

\newcommand\ALG@tikzmark@start{%
    \global\let\ALG@tikzmark@last\ALG@tikzmark@starttext%
    \expandafter\edef\csname ALG@tikzmark@\theALG@nested\endcsname{\theALG@tikzmark@tempcnta}%
    \tikzmark{ALG@tikzmark@start@\csname ALG@tikzmark@\theALG@nested\endcsname}%
    \addtocounter{ALG@tikzmark@tempcnta}{1}%
}

\def\ALG@tikzmark@starttext{start}
\newcommand\ALG@tikzmark@end{%
    \ifx\ALG@tikzmark@last\ALG@tikzmark@starttext
    \else
        \tikzmark{ALG@tikzmark@end@\csname ALG@tikzmark@\theALG@nested\endcsname}%
        \tikz[overlay,remember picture] \draw[\ALGtikzmarkcolor] let \p{S}=($(pic cs:ALG@tikzmark@start@\csname ALG@tikzmark@\theALG@nested\endcsname)+(\ALGtikzmarkextraindent,\ALGtikzmarkverticaloffsetstart)$), \p{E}=($(pic cs:ALG@tikzmark@end@\csname ALG@tikzmark@\theALG@nested\endcsname)+(\ALGtikzmarkextraindent,\ALGtikzmarkverticaloffsetend)$) in (\x{S},\y{S})--(\x{S},\y{E});%
    \fi
    \gdef\ALG@tikzmark@last{end}%
}

\apptocmd{\ALG@beginblock}{\ALG@tikzmark@start}{}{\errmessage{failed to patch}}
\pretocmd{\ALG@endblock}{\ALG@tikzmark@end}{}{\errmessage{failed to patch}}
\makeatother

\title{Contrastive Learning in Memristor-based Neuromorphic Systems}

\author{%
Cory Merkel\\
Brain Lab \\
Rochester Institute of Technology \\ 
Rochester, NY 14623\\
\texttt{cemeec@rit.edu}
\And
Alexander G. Ororbia\\
The Neural Adaptive Computing Laboratory \\
Rochester Institute of Technology \\
Rochester, NY 14623\\
\texttt{ago@cs.rit.edu}
}

\begin{document}

\maketitle
\begin{abstract}
Spiking neural networks, the third generation of artificial neural networks, have become an important family of neuron-based models that sidestep many of the key limitations facing modern-day backpropagation-trained deep networks, including their high energy inefficiency and long-criticized biological implausibility. In this work, we design and investigate a proof-of-concept instantiation of contrastive-signal-dependent plasticity (CSDP), a neuromorphic form of forward-forward-based, backpropagation-free learning. Our experimental simulations demonstrate that a hardware implementation of CSDP is capable of learning simple logic functions without the need to resort to complex gradient calculations.
\end{abstract}

\section{Introduction}
\label{sec:intro}

Neuromorphic computing based on low-power memristors is receiving significant attention lately due to the large energy and compute requirements of deep learning models \cite{strubell2020energy,hendy2022review}. However, on-chip training of neuromorphic systems remains a significant challenge due to the hardware complexity of backpropagation of errors (BP) and its approximations.  In this paper, we report preliminary work on the design of BP-free training circuitry for memristor-based neuromorphic systems that follows the contrastive learning (CL) paradigm \cite{le2020contrastive,jaiswal2020survey}.  CL builds latent representations of training data by learning differences between positive and negative examples.

\begin{figure}[!t]
\centering
\includegraphics[width=0.575\columnwidth]{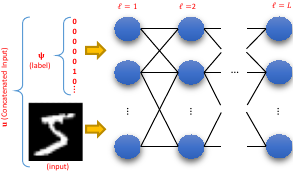}
\vspace{-2mm}\caption{Neuronal network setup for the FF/CSDP algorithm.}
\label{fig:ff}
\vspace{-6mm}
\end{figure}

The procedure recently proposed in \cite{hinton2022forward} -- the forward-forward (FF) CL algorithm -- is especially attractive because its greedy layer-wise weight updates remove the need for backpropagation, thereby circumventing many of its inherent issues and biological implausibilities \cite{crick1989recent,ororbia2023brain}, such as the weight transport problem \cite{grossberg_resonance_1987}.  However, while many developments have been made to advance the mechanics of forward-forward learning \cite{ahamed2023forward,giampaolo2023investigating,terres2024improvement,aminifar2024lightweight} as well as to improve certain aspects of its biological plausibility \cite{ororbia2023predictive}, the algorithm has yet to be demonstrated on a neuromorphic hardware platform.  
Figure \ref{fig:ff} shows the basic setup for an FF-trained neural network used in a supervised learning task, i.e., predicting some target vector given a (sensory) input vector.  In this case, a fully-connected feedforward neural network will take in training example patterns and their corresponding labels as inputs.  The concatenation of these two inputs makes up one `positive' example.  FF attempts to find synaptic weight values that yield high network activity for positive examples and low network activity for `negative' examples.  One simple and effective way to create negative examples, for supervised learning setups, is to pair training input examples with incorrect labels.  Note that this objective can be accomplished in a layer-wise fashion that does not depend on the global, forward and backward-locked feedback pathway induced by backpropagation \cite{jaderberg2017decoupled,ororbia2023brain}.  

Recently, a generalized version of the FF algorithm, called contrastive-signal-dependent plasticity (CSDP) \cite{ororbia2023learning}, was proposed.  CSDP can desirably compute synaptic weight updates based on spike train statistics, e.g. spike traces.  In this research, we designed a proof-of-concept implementation of CSDP in $45$ nm CMOS integrated with a semi-empirical memristor model.  Leaky integrate-and-fire (LIF) neurons with a spike rate encoding were outfitted with trace circuits.  The trace circuits maintain a running average of the neuron's spike rate represented as voltage on a capacitor.  Then, a simple voltage-to-time converter is used to generate a pulse width representation of the trace. 
For positive examples, synaptic weight strengths with large pre- and post-synaptic spike traces are increased, and for negative examples, they are decreased.

Our preliminary results indicate that the hardware implementation of CSDP can learn a simple logic function in both single and multi-layer networks without the need for loss gradient calculations over multiple layers (such as those produced by BP).  We remark that future work should focus on comparisons between CSDP and other hardware-friendly training algorithms such as spike timing dependent plasticity (STDP) \cite{markram2011history}, random feedback alignment \cite{samadi2017deep}, and surrogate-based (approximate BP) approaches \cite{yin2017algorithm}. 
The central contributions made in this paper are as follows:
\begin{itemize}[noitemsep,nolistsep]
    \item We implement the generalization of FF for spiking neuronal dynamics, known as CSDP, for memristor-based neuromorphic platforms; 
    \item We provide positive, proof-of-concept results related to the performance of our neuromorphic instantiation of CSDP-based credit assignment.
\end{itemize}

\section{Forward-Forward and Contrastive-Signal-Dependent Plasticity Preliminaries}
\label{sec:forward_forward_csdp}

\subsection{The Forward-Forward Algorithm}
\label{sec:ff_algo}

Suppose we have a set of positive training examples $\mathcal{D}^{+}$ and a corresponding set of negative examples $\mathcal{D}^{-}$ that we would like to differentiate between.  Then, optimization in the context of FF learning can be formally described in the following manner:
\begin{equation}
\mathbf{W}^{(\ell)}=\underset{\mathbf{W}}{\arg\max}\:\mathbb{E}[g^{(\ell)}(\mathcal{D}^{+};\mathbf{W})-g^{(\ell)}(\mathcal{D}^{-};\mathbf{W})] \label{eqn:ff_rule}
\end{equation}
where $\mathbf{W}^{(\ell)}$ is the synaptic weight matrix for layer $\ell$, and $g^{(\ell)}(\mathcal{D}^{+};\mathbf{W}^{(\ell)})$ and $g^{(l)}(\mathcal{D}^{-};\mathbf{W}^{(\ell)})$ are the \textit{goodness} measures, with respect to the positive $\mathcal{D}^{+}$ and negative examples $\mathcal{D}^{-}$ (for neuronal layer $\ell$), respectively.  Goodness is effectively a measurement of a neuronal layer's (overall) activity, with local cost functionals that are designed to measure quantities such as the length of the activity vector (e.g., the L$2$ Euclidean distance). Then, goodness values that are above a particular threshold $\theta$ correspond to inputs from $\mathcal{D}^{+}$, while goodness values below $\theta$ correspond to inputs from $\mathcal{D}^{-}$.  Therefore, the probability that a particular input is positive can be calculated as follows: 
\begin{equation}
p^{(\ell)}(\mathbf{u}\in \mathcal{D}^{+})=\sigma(g^{(\ell)}\left(\mathbf{u}\right)-\theta)
\end{equation}
where $\sigma$ is the logistic sigmoid function.  Using a cross-entropy loss, the gradient descent weight update can then be derived as: 
\begin{equation}
\Delta w_{ij}^{(\ell)}=-\alpha\epsilon\frac{\partial g^{(\ell)}(\mathbf{u})}{\partial x_{i}^{(\ell)}}f'(s_{i}^{(\ell)})x_{j}^{(\ell)},
\end{equation}
where $\alpha$ is the learning rate, $x_{i}^{(\ell)}$ is the output of neuron $i$ in layer $\ell$, $s_{i}^{(\ell)}$ is the input of neuron $i$ in layer $\ell$, the error term is $\epsilon\equiv\sigma(g^{(\ell)}\left(\mathbf{u}\right)-\theta)-y$, $y$ is 0 for negative examples and $1$ for positive examples, and $f'$ is the derivative of a neuron's activation function. If $g$ is the squared L2 norm of a layer's output, and $f$ is the rectified linear unit function, then the synaptic weight update becomes an error-modulated Hebbian learning rule:
\begin{equation}
\Delta w_{ij}^{(\ell)}=-2\alpha\epsilon x_{i}^{(\ell)}x_{j}^{(\ell-1)}.
\end{equation}
Note, in \cite{hinton2022forward}, the full network is typically trained layer-by-layer (greedy layer-wise), from input to output, each time minimizing the cross-entropy loss with all previous layer weights frozen.  After training, inference can then be performed by providing an input to the network without a label and then searching through the label space in order to find the input/label pair that produces the largest cumulative goodness computed over all layers:
\vspace{-2mm}
\begin{equation}\vspace{-2mm}
    \hat{\psi}^{(p)}=\underset{\mathbf{\psi}}{\arg\max}\:\sum\limits_{\ell} g^{(\ell)}(\mathbf{u}^{(p)})
\end{equation}
where the superscript $p$ specifies the particular input and $\psi$ is the label.


\subsection{Contrastive-Signal-Dependent Plasticity}

The generalization of FF learning that this study will adapt and optimize for neuromorphic platforms is referred to as contrastive-signal-dependent plasticity (CSDP) \cite{ororbia2023learning}. CSDP generalizes the FF update rule -- the derivative of Equation \ref{eqn:ff_rule} with respect to $\mathbf{W}^{(\ell)}$ -- to spike trains by formulating the FF cost functional as a modulatory signal $\delta^{(\ell)}$ applied to post-synaptic traces that are maintained by neuronal cells within layer $\ell$ of a spiking neuronal system.

This work makes use of leaky integrate-and-fire (LIF) neurons that exhibit the following behavior with respect to their membrane potential(s) $v_{m_{i}}^{(\ell)}$: 
\begin{equation}
C_{m}\frac{\mathrm{d}v_{m_{i}}^{(\ell)}(t)}{\mathrm{d}t}=\mathbf{1}_{s_{i}^{(\ell)}>0}(I_{in}-I_{l})
\end{equation}
where $C_{m}$ is the membrane capacitance, $I_{in}$ is the neuron's incoming current from pre-synaptic activity, and $\mathbf{1}$ is the indicator function. Notice that we use a variant of the LIF neuron with a constant leak factor ($I_{l}$). When the neuron's membrane potential reaches a threshold $\phi$, it will produce a spike and the membrane is reset to zero.  An absolute refractory period dictates the minimum time between spikes (maximum spiking frequency), which we call $T_{r}$.  While spiking neurons may employ a number of encoding strategies \cite{paugam2012computing}, here we consider simple rate encoding where the number of spikes in a specified time window encodes the neuron output. The spike rate for a constant input current is:
\begin{equation}
\nu_{i}^{(\ell)}=\frac{1}{\frac{\phi C_{m}}{I_{in_{i}}-I_{l}}+T_{r}+T_{w}}
\label{eqn:spikerate}
\end{equation}
where $T_{w}$ is the width of the spike.  Note that software implementations of SNNs usually assume that the spike is a delta function, so $T_{w}=0$.  However, in this work, our hardware neuron design has a non-zero spike width.  Neurons in the input layer will be connected to a constant current source, but deeper layers of the network will have connections to other spiking neurons.  Each time a pre-synaptic neuron spikes, it injects a current into the post-synaptic neuron that is proportional to the synaptic weight $w_{ij}^{(\ell)}$.  A simple, closed-form model of the post-synaptic spike rate as a function of the pre-synaptic spike rates is difficult to express because, although information is carried in the spike rates, the relative timing of pre-synaptic spikes also has an impact on how the membrane potential changes.  In this work, we will model the approximate response of the post-synaptic neuron to pre-synaptic spikes by considering the average current flowing into and out of the post-synaptic membrane over time. This can be formally expressed as follows:
\begin{equation}
\bar{I}_{in_{i}}=\sum\limits_{j}\nu_{j}^{(\ell-1)}T_{w}w_{ij}^{(\ell)}I_{s}-I_{l} 
\end{equation}
where $I_{s}$ is a constant that converts the spike count multiplied by the weight into a current.  In addition, the spike count in a given time window $T$ is: 
\begin{equation}
\eta_{i}^{(\ell)} = \lfloor (T+T_{r})\nu_{i}^{(l)} \rfloor . 
\label{eqn:spikecount}
\end{equation}
Note that if $T$ is large enough to capture low spike rates of interest, then the spike count provides a good/useful proxy for the spike rate.  

The aforementioned modulatory signal $\delta^{(\ell)}$ of CSDP is used to construct a synaptic update rule that centers around a modulated contrastive pairing of two Hebbian adjustments; effectively a pairing of two goodness-modulated pre-synaptic spike-timing-dependent plasticity (STDP) rules. As a result, for any synaptic weight $w^{(\ell)}_{ij}$ connecting pre-synaptic neuronal cell $j$ in region/layer $\ell-1$ 
to post-synaptic neuronal cell $i$ in region/layer $\ell$, 
the update to synaptic efficacy takes on the following form:
\begin{equation}
\begin{split}
    \Delta w^{(\ell)}_{ij} &\propto \tau_w \frac{\partial w^{(\ell)}_{ij}}{\partial t}\\
    &= \gamma^{(+)}\delta^{(\ell,+)} (z^{(\ell,+)}_i o^{(\ell-1,+)}_j) \\
    &- \gamma^{(-)}\delta^{(\ell,-)} (z^{(\ell,-)}_i o^{(\ell-1,-)}_j) \label{eqn:csdp_rule}
\end{split}
\end{equation}
where $o$ is a binary variable indicating the presence (one) or absence (zero) of a spike pulse; for instance, $o^{(\ell-1)}_j$ specifically indicates if the pre-synaptic neuron $j$ in layer $(\ell-1)$ has emitted a pulse or not. $\tau_w$ is the synaptic update time constant, 
$\delta^{(\ell,+)}$ is the positive modulator signal, weighted by the scale factor $\gamma^{(+)} \in [0,1]$), and $\delta^{(\ell,-)}$ is the negative modulator signal, weighted by the scale factor $\gamma^{(-)} \in [0,1]$). Crucially, $z^{(\ell,+)}_i$ is the trace of the (post-synaptic) spiking $i$-th neuronal cell in layer $\ell$. Note that, for the modulator signal $\delta$, trace value $z$, and spike pulse variable $o$, we attach the superscript ``$+$'' to denotes a statistic in the positive phase (corresponding to an input data pattern sampled from $\mathcal{D}^{+}$) while ``$-$'' denotes a negative phase statistic (corresponding to an input pattern sampled from $\mathcal{D}^{-}$). 

For Equation \ref{eqn:csdp_rule}, the trace value for a spiking neuronal cell adheres to the following differential equation: 
\begin{align}
    \frac{\partial z^{(\ell,+)}_i }{\partial t} = -\frac{z^{(\ell,+)}_i}{\tau_z} + \gamma_z o^{(\ell,+)}_i  \label{eqn:csdp_trace}
\end{align}
where $\tau_z$ is the trace time constant (in ms) and $\gamma_z$ is a non-negative scalar that controls the magnitude of the increment applied to the trace upon the occurrence of a spike pulse $o$ (we remark that setting $\gamma_z = 1$ and $\tau_z = \infty$ results in a form of spike counting). Note that Equation \ref{eqn:csdp_trace} depicts a trace being maintained for positive-phase spikes; however, the equation is the same for negative-phase spikes, simply requiring the replacement of the ``$+$'' symbol with the `$-$' symbol. 

Finally, it is important to point out that although Equation \ref{eqn:csdp_rule} depicts CSDP as applying two simultaneous, parallel pre-synaptic-driven\footnote{Note that this phrase means that the pre-synaptic pulse events trigger updates.} STDP updates for both positive and negative data, one does not necessarily need to strictly follow this format. For instance, if the system is only provided with positive data samples, then only the STDP term modulated by $\delta^{(\ell,+)}$ would be applied (during the time period where only positive examples are presented). If, on the other hand, the system is only given negative data samples, then only the STDP term modulated by $\delta^{(\ell,-)}$ would be triggered. This would allow interleaving of positive and negative samples (and even periods of time where only positive examples are presented followed by periods of only negative examples) and usefully permit sequential presentation of positive and/or negative data points (and thus not require running two parallel instances of the spiking neuronal network).

\section{Hardware Building Blocks}
\label{sec:hardare}

This section describes the design of the hardware building blocks for implementing the CSDP algorithm in memristor-based neuromorphic hardware.

\subsection{CMOS and Memristor Technology}
\label{sec:cmos_memristor}

\begin{figure}[!t]
    \centering
    \includegraphics[width=0.575\linewidth]{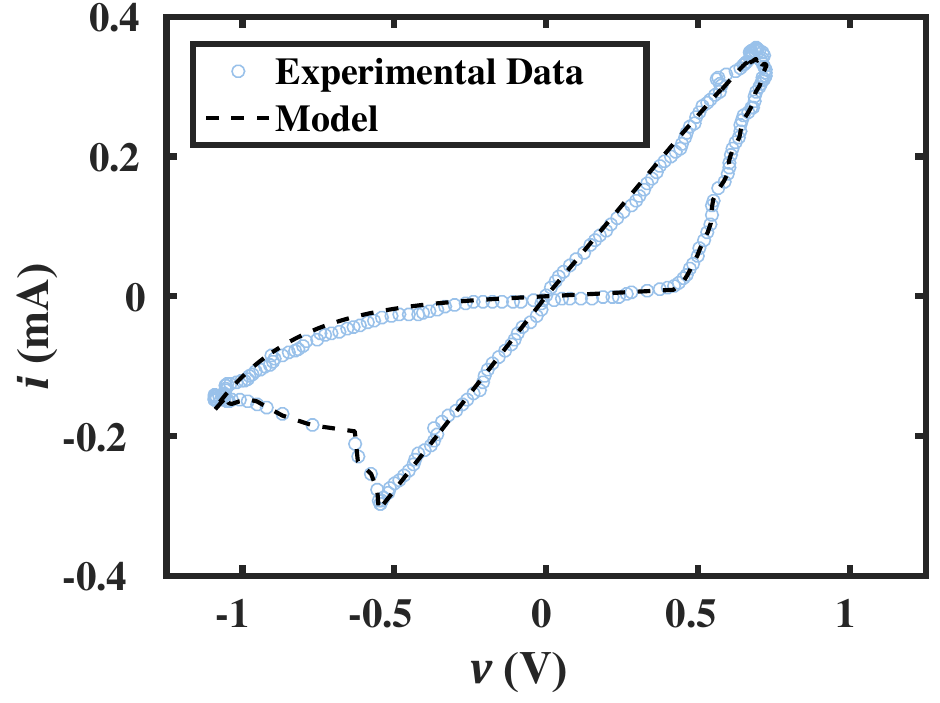}
    \caption{Memristor current-voltage relationship and model fit.}
    \label{fig:memiv}
\vspace{-4mm}
\end{figure}

All of our hardware block designs and simulations are completed using LTspice \cite{ltspice}.  We used a $45$ nm high performance predictive technology MOSFET model \cite{ptm} with a nominal $1$ V power supply for all CMOS portions of the design.  In addition to MOSFETs, our synapse design also makes use of memristors \cite{chua1971memristor,strukov2008missing}, which can be described as non-volatile electronically-tunable resistors.  A semi-empirical model, developed by Merkel in \cite{merkel2015design}, was used in the synapse circuit design and analysis.  The I-V characteristics when the device is not switching can be described as:
\begin{equation}
i=\left\{\begin{array}{cl}
\chi G_{mon}v+\left(1-\chi\right)G_{off}\xi_{1}^{+}\mathrm{sinh}\left(\frac{v}{\xi_{1}^{+}}\right), v\ge 0\\
\chi G_{on}v+\left(1-\chi\right)G_{off}\xi_{1}^{-}\mathrm{sinh}\left(\frac{v}{\xi_{1}^{-}}\right), v< 0
\end{array}\right.,
\end{equation}
where $i$ and $v$ are the current through and the voltage across the memristor, $\chi\in[0,1]$ is the memristor state variable, $G_{off}$ and $G_{on}$ are the off and on conductance values corresponding to $\chi=0$ and $\chi=1$, and $\xi_{1}^{+(-)}$ are fitting parameters for the positive (negative) portion of the I-V characteristic.  Notice that, when $v$ is small, $\mathrm{sinh}\left(v/\xi_{1}^{+(-)}\right)\approx v/\xi_{1}^{+(-)}$, thus: 
\begin{equation}
\begin{array}{cr}
i\approx \left[\chi G_{on}+\left(1-\chi\right)G_{off}\right]v, & v \approx 0,
\end{array}
\end{equation}
which provides a linear interpolation between the extreme conductance values based on the state $\chi$.  When the applied voltage is above or below particular threshold values, the memristor's state will change, with a positive voltage increasing $\chi$ and a negative voltage decreasing $\chi$.  The dynamics of the memristor state change -- which will be related to the dynamics of the neural network synaptic weight modification -- are generally non-linear and dependent on the exact physical switching mechanism at play. Our semi-empirical model, based on common device physics of the switching behavior, has flexibility to fit the non-linear nature of the switching \cite{merkel2015design}.  For simplicity, however, we will  assume that the switching behavior is an approximately linear function of the applied flux linkage (a product of the applied voltage and time), resulting in the following: 
\begin{equation}
\mathrm{d}\chi \propto \left\{\begin{array}{cl} H(v-V_{tp})(v-V_{tp})\mathrm{d}t, v>0 \\
H(V_{tn}-v)(V_{tn}-v)\mathrm{d}t, v<0 
\end{array}\right.
\end{equation}
where $H(\cdot)$ is the Heaviside step function, $V_{tp}$ is the positive threshold voltage for memristor switching, and $V_{tn}$ is the negative threshold for memristor switching.  Note that $\chi$ has a hard bound between $0$ and $1$.  In this work, we fit our model to the experimental memristor data from \cite{Oblea2010}. Figure \ref{fig:memiv} shows the memristor experimental data as well as our model fit.

\begin{table}[]
\centering
\caption{Memristor model parameters.}
\label{tab:memristor}
\begin{tabular}{cccc}
\toprule
 $G_{on}$ & $G_{off}$ & $V_{tp}$ & $V_{tn}$ \\
 \midrule \vspace{-2mm}\\
 (1800 $\Omega$)$^{-1}$ & (4.637$\times 10^{4}$ $\Omega$)$^{-1}$ & 0.4 V & -0.55 V  \\
 \bottomrule
\end{tabular}
\end{table}

\subsection{LIF Neuron Design}
\label{sec:lif_hardware}

\begin{figure}[!t]
\centering
\subfigure[]{
\raisebox{5mm}{
\includegraphics[width=0.6\columnwidth]{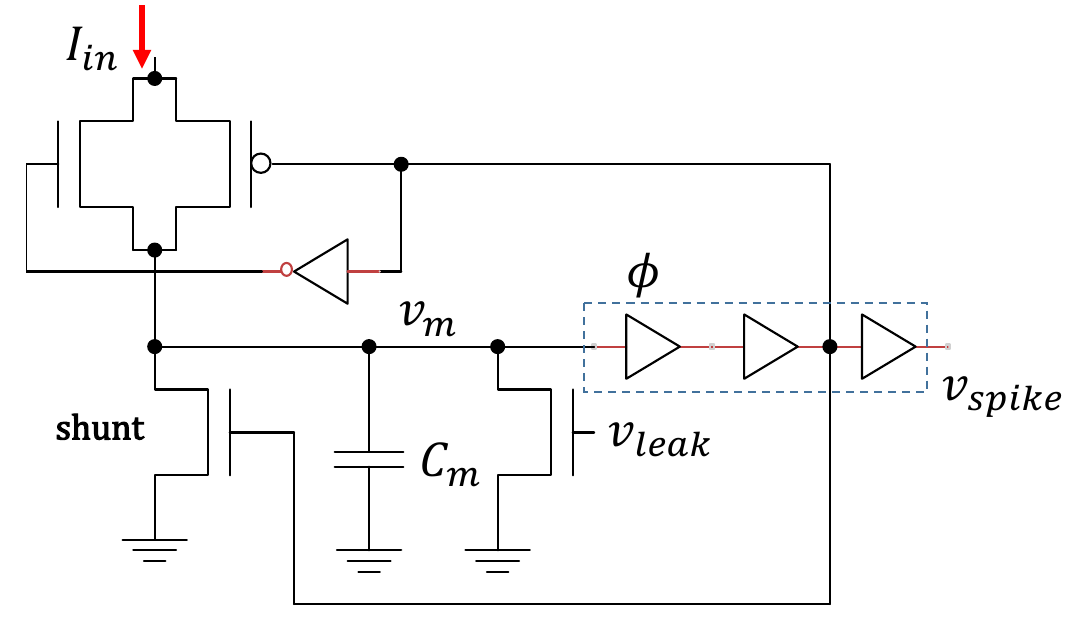}}
}
\hspace{-4mm}
\subfigure[]{

\includegraphics[width=0.35\columnwidth]{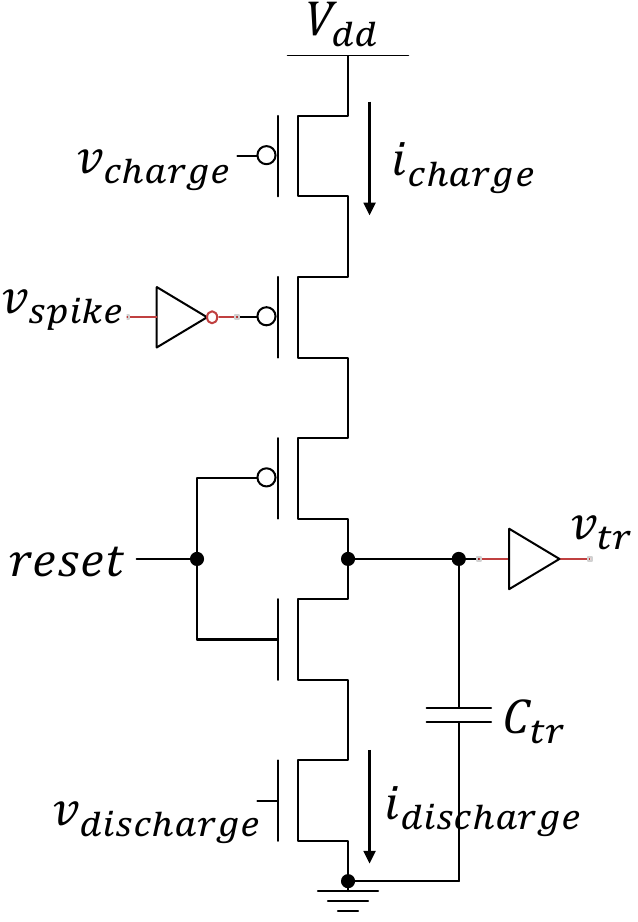}}
\caption{(a) LIF neuron design. (b) The circuit designed to keep track of a spiking neuron's trace.}
\label{fig:simplelif}
\end{figure}

Several neuromorphic neuron designs have been proposed in the past to model varying degrees of biologically-realistic behavior \cite{indiveri2011neuromorphic}.  In this work, we use a very simple LIF neuron design that encodes information using spike rate, has a constant leak, and has an absolute refractory period. The design for this is shown in Figure \ref{fig:simplelif}(a). Input current flows into the neuron through a transmission gate that is turned on in the resting state, when $v_{m}=0$.  As the membrane capacitor charges from $I_{in}$, an NMOS transistor with a constant reference voltage $v_{leak}$ serves to provide an approximately constant leak current.  When the membrane potential reaches the threshold of the buffer chain, the output $v_{spike}$ goes from $0$ to $V_{dd}$.  This also causes the transmission gate to turn off, disconnecting the input current from the membrane capacitor. At the same time, an NMOS transistor (labeled shunt) is turned on to quickly reset the membrane potential back to $0$. From this point, the delay of the signal through the two buffers back to the transmission gate and shunt transistor defines the absolute refractory period. Note that we use an additional buffer before the outgoing spike signal in order to isolate the feedback dynamics from the load that the neuron itself will ultimately drive.

We utilized $v_{leak}=0.5$ V, giving rise to a constant leak current around $365$ nA.  The membrane capacitance was set at $C_{m} = 120$ fF.  The spike width and refractory period were measured to be $13$ ns and $12$ ns, respectively, and the threshold of the buffer chain was measured to be around $700$ mV. These values are summarized in Table \ref{tab:neuron}.  All transistor lengths were set to the minimum size of $45$ nm, with the exception of the devices in the inverter that controls the transmission gate and the first buffer in the buffer chain. These were sized with $4$$\times$ the minimum length in order to achieve high gain (steep rise and fall transitions) for the spike.  The length of the leak transistor was also sized at $4$$\times$ the minimum length in order to achieve the desired leak rate. All widths were sized at $45$ nm except for the shunt transistor and the transmission gate transistors, which had $4$$\times$ widths.

\begin{figure}[!t]
\centering
\includegraphics[width=0.65\columnwidth]{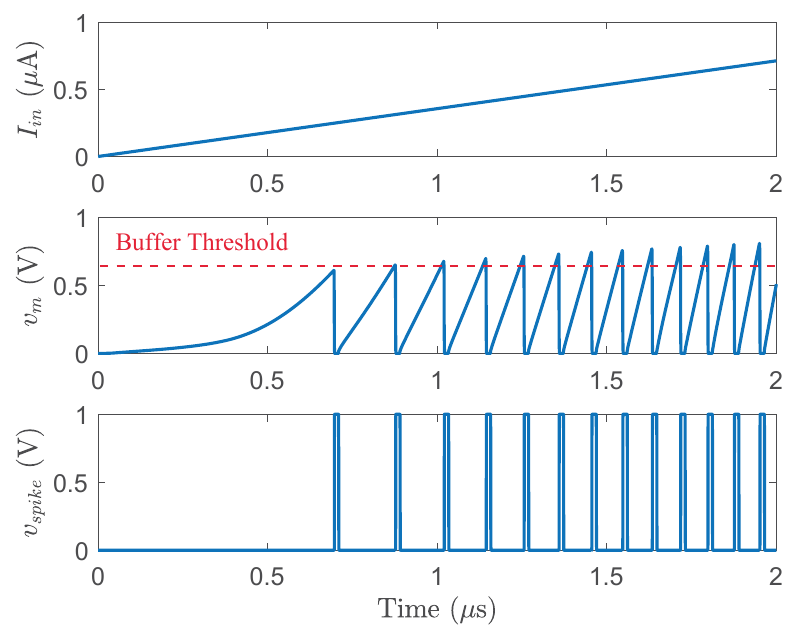}
\caption{LIF neuron behavior. \emph{Top}: Input current versus time. \emph{Middle}: Membrane potential versus time. \emph{Bottom}:  Neuron output versus time.}
\label{fig:simplelifwaveform}
\vspace{-4mm}
\end{figure}

\begin{figure}[!t]
\centering
\includegraphics[width=0.55\columnwidth]{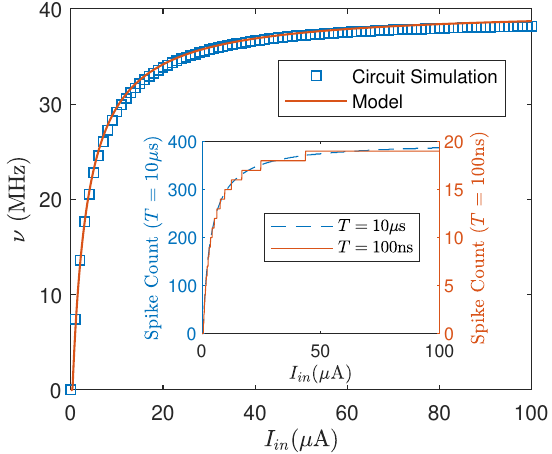}
\caption{Spike rate and spike count vs. neuron input current.}
\label{fig:simplelifresults}
\end{figure}

The neuron behavior is shown in Figure \ref{fig:simplelifwaveform} for a ramping input current.  The spike frequency increases as a result of the increased stimulus.  Note that for smaller input currents, the membrane voltage does not rise above the buffer threshold before it is reset to zero.  However, as the input current becomes larger, the delay in the reset mechanism allows for the membrane voltage to overshoot the neuronal firing threshold. Although this deviates from the ideal LIF behavior, we have not observed this to have any significant impact on the neuron spike output. The spike frequency of the neuron versus constant input current is further shown in Figure \ref{fig:simplelifresults}. Here, we see that the neuron spike rate saturates around $\nu_{max} = 40$ MHz, which is in line with the limit of (Equation \ref{eqn:spikerate}) as $I_{in}$ becomes very large.  In fact, the model of the spike rate in (Equation \ref{eqn:spikerate}) is in almost perfect agreement with the observed circuit simulation data.

\begin{table}[]
\centering
\caption{Spiking neuron hardware parameters and values.}
\label{tab:neuron}
\begin{tabular}{ccccc}
\toprule
 $v_{leak}$ & $I_{l}$ & $T_{r}$ & $T_{w}$ & $C_{m}$ \\
 \midrule \vspace{-2mm}\\
 0.5 V & 365 $\mu$A & 12 ns & 13 ns & 120 fF    \\
 \bottomrule
\end{tabular}
\vspace{-4mm}
\end{table}

\subsection{Trace Circuit}
\label{sec:trace_circuit}

A simple trace circuit was designed to keep track of the neuron spike rate over windows of length $T$. This circuit is shown in Figure \ref{fig:simplelif}(b). Each spiking neuron has its own associated trace circuit. When $reset=0$, each time the neuron spikes, charge is added to the storage capacitor $C_{tr}$, causing an increase in the capacitor voltage of $T_{w}i_{charge}/C_{tr}$. When $reset=1$, the capacitor will decrease at a rate of $i_{discharge}/C_{tr}$. Notice that there is no leakage, so our particular trace implementation essentially tracks the spike count over whatever window size we choose, corresponding to $\tau_{z}=\infty$. The voltages $v_{charge}$ and $v_{discharge}$ are constant reference voltages set to achieve the desired $i_{charge}$ and $i_{discharge}$. The current $i_{charge}$ is set such that the trace capacitor charges to $V_{dd}$ when the neuron is firing at its maximum spike rate over the time window $T$.  This leads to the following relationship:
\begin{equation}
i_{charge}=\frac{V_{dd}C_{tr}}{\nu_{max}TT_{w}} . 
\label{eqn:icharge}
\end{equation}
For training, it will be convenient to represent the trace value as a digital pulse width rather than an analog voltage. This conversion is achieved by adding a buffer at the output of the trace circuit and controlling the discharge rate of $C_{tr}$ by $i_{discharge}$.

\begin{figure}[!t]
\centering
\includegraphics[width=0.64\columnwidth]{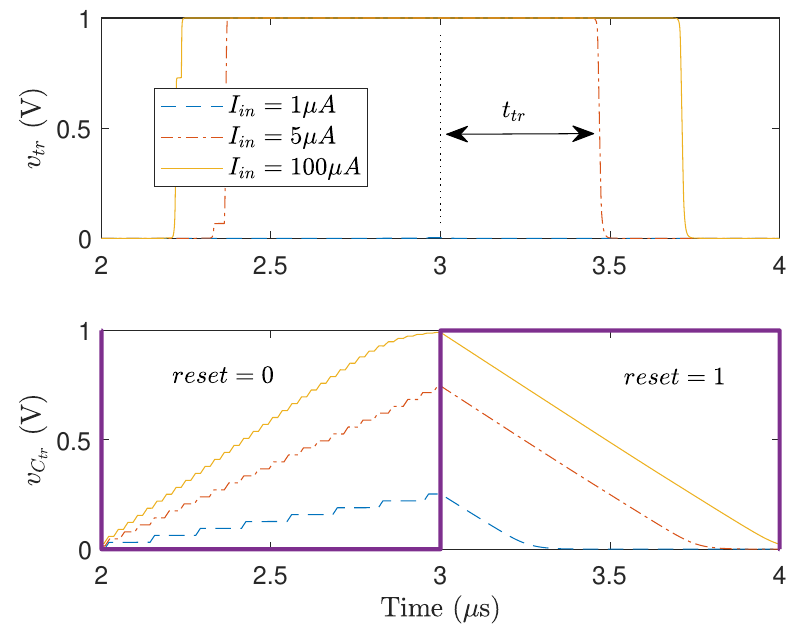}
\caption{Trace circuit behavior for different neuron input currents.}
\label{fig:tracecircuitwave}
\vspace{-5mm}
\end{figure}

An example of the trace circuit operation is shown in Figure \ref{fig:tracecircuitwave}. In this case, we have used $T=1 \mu s$, $i_{charge} = 240$ nA, $i_{discharge} = 100$ nA, and $C_{tr} = 100$ fF.  Note that $i_{charge}$ is set to a slightly larger value than dictated by (Equation \ref{eqn:icharge}) to account for non-idealities such as parasitic capacitances.  In the example in Figure \ref{fig:tracecircuitwave}, three different neuron input currents are used to drive the neuron that connects to the trace circuit. In the case of $I_{in}=$1 $\mu$A, the voltage on the trace capacitor, $v_{C_{tr}}$ does not integrate to a value above the buffer threshold during $T$, causing no pulse to be produced when the $reset$ becomes $1$.  As the stimulating input current becomes larger, $v_{C_{tr}}$ reaches larger values, causing larger pulse widths at the output of the trace circuit.

\subsection{Memristor Synapse Design}
\label{sec:memristor_synapses}

\begin{figure}[!t]
\centering
\subfigure[]{\raisebox{5mm}{\includegraphics[width=0.28\columnwidth]{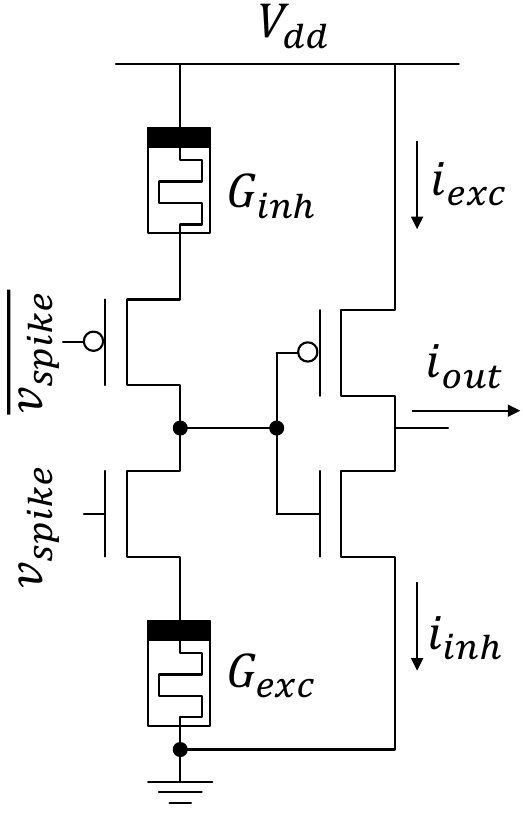}}}
\hspace{5mm}
\subfigure[\hspace{-8mm}]{\includegraphics[width=0.38\columnwidth]{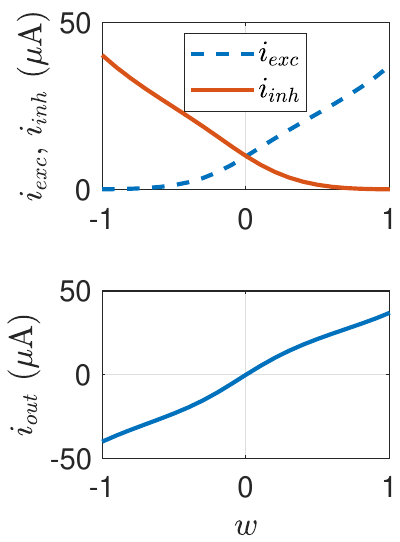}}
\caption{(a) Memristor synapse design used in this work and (b) its transfer characteristics.}
\label{fig:synapse}
\vspace{-5mm}
\end{figure}

In this work, we designed a novel memristor synapse circuit that has good dynamic range and linearity, even for relatively small ranges of memristor conductance values. The design is shown in Figure \ref{fig:synapse}(a). An incoming spike turns on the NMOS and PMOS transistor, which allows current to flow in the left branch, giving rise to a voltage that drives two opposing current sources in the right branch. The voltage that divides the current sources depends on the ratio of conductances $G_{inh}$ to $G_{exc}$. Note that a larger inhibitory conductance $G_{inh}$ increases the driving voltage, causing larger inhibitory current, $i_{inh}$; on the other hand, a larger excitatory conductance $G_{exc}$ reduces the driving voltage, causing a larger excitatory conductance $i_{exc}$. The final current output is $i_{out}=i_{exc}-i_{inh}$.  The synapse behavior is shown in Figure \ref{fig:synapse}(b). We define the weight $w$ according to a linear interpolation of each memristor between its on and off conductance.  That is, when $w=-1$, $G_{inh}=G_{on}$ and $G_{exc}=G_{off}$ and when $w=1$, $G_{inh}=G_{off}$ and $G_{exc}=G_{on}$.  Not shown in the synapse schematic are the access switches that will be present at the positive and negative terminals of both memristors to provide programming voltages during training.

\subsection{Training Circuit}
\begin{figure}[!t]
\centering
\includegraphics[width=0.315\columnwidth]{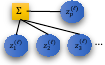}
\caption{Computation of $p^{(\ell)}(\mathbf{u}\in\mathcal{D}^{+})$ as a trace.}
\label{fig:tracecsdp}
\end{figure}

We implement a purely trace-based version of CSDP, where the goodness as well as the probability that an input is a positive example are computed by connecting all of the neurons within a layer to an additional spiking neuron using constant positive weights (in \cite{ororbia2023learning}, this neuronal unit was biologically interpreted as a simple type of astrocytic support cell). The scheme is shown in Figure \ref{fig:tracecsdp}. The trace of the additional neuron will reflect the sum of the spike rates of all of the neurons within the layer, which should be large for positive input examples and small for negative input examples.  Now, our trace-based CSDP update rule is formally:
\begin{equation}
\begin{split}
\Delta w_{ij}^{(\ell)}&=-\alpha\left(\frac{z_{p}^{(\ell)}}{z_{max}}-y\right)z_{i}^{(\ell)}z_{j}^{(\ell-1)}=-\alpha\epsilon^{(\ell)}z_{i}^{(\ell)}z_{j}^{(\ell-1)}\\
&\approx-\alpha\mathrm{sgn}(\epsilon^{(\ell)})\mathrm{min}(|\epsilon^{(\ell)}|,z_{i}^{(\ell)},z_{j}^{(\ell-1)})
\end{split}
\end{equation}
where $z_{max}$ is the maximum trace value, and the final approximation makes use of the fact that the magnitude of the product of two numbers with magnitudes between $0$ and $1$ can be roughly estimated as their minimum value \cite{merkel2017neuromemristive}, which can be computed using a simple AND gate connected to the pulse representations of each trace. This gives an output voltage pulse representation of the minimum value that we use to control the amount of time to apply a positive (in the case that $\mathrm{sgn}(\epsilon^{(\ell)})$ is negative) or a negative (in the case that $\mathrm{sgn}(\epsilon^{(\ell)})$ is positive) programming voltage to the associated memristors.

\section{Logic Function Example}
\label{sec:toy_logic_task}

\begin{figure}[!t]
\centering
\subfigure[]{
\includegraphics[width=0.605\columnwidth]{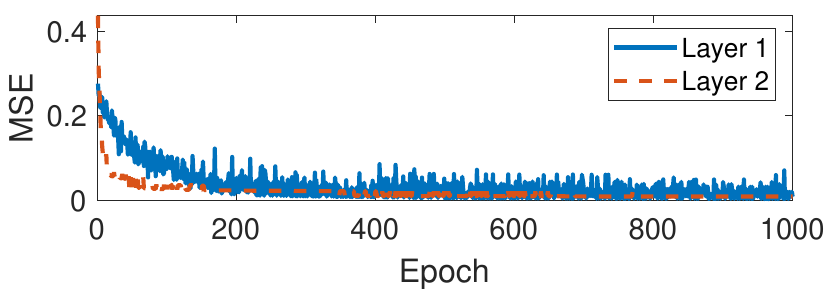}
}
\subfigure[]{
\includegraphics[width=0.8\columnwidth]{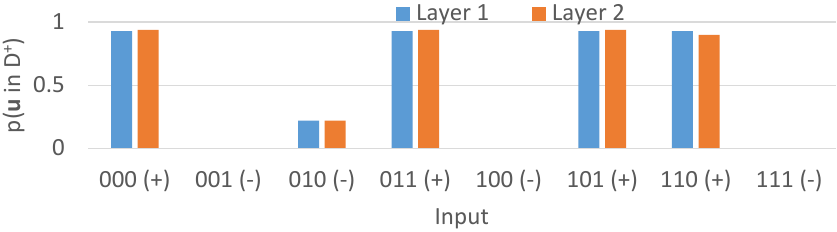}
}
\caption{Results for a 2-layer neural network trained with our neuromorphic instantiation of CSDP. (a) The MSE of each layer over training epoch. 
(b) Each layer's final calculated probability that an input (pattern) is positive.}
\label{fig:xorresults}
\vspace{-3mm}
\end{figure}

As a proof-of-concept, we built a behavioral model of our hardware in MATLAB and simulated it to solve the XOR problem. The behavioral model captures all of the key characteristics of the building blocks discussed in the last section (maximum spike rates, synaptic weight ranges, etc.). The simple network has 3 inputs, two of which are the logical inputs, and one which is the label input.  It also has 2 layers with $5$ and $3$ neurons, respectively.  Figure \ref{fig:xorresults} shows a typical set of results. In Figure \ref{fig:xorresults}(a), we show the MSE of each layer's probability output, $p^{(\ell)}(\mathbf{u}\in\mathcal{D}^{+})$, versus training epoch. Observe that both neuronal layers converge to small MSE loss values.  Figure \ref{fig:xorresults}(b) shows the performance of the model after training. Each group of bars in the plot corresponds to the total concatenated input and whether the input is positive or negative.  For example, ``001(-)" means that the supplied input to the XOR gate is ``00" and the expected output is `1', which is obviously a negative example since the output should be `0'.  We observe that all of the positive inputs lead to high probabilities and all of the negative inputs lead to low probabilities, giving 100\% accuracy.

\section{Conclusions}
\label{sec:conclusion}

In this work, we designed and studied a proof-of-concept neuromorphic implementation of contrastive-signal-dependent plasticity (CSDP), a generalization of forward-forward contrastive learning to the case of spiking neuronal dynamics. We further analyzed our supporting circuit/component hardware designs to ensure that they exhibited properties/behavior that would be conducive to resulting in and generating the necessary measurement values needed to calculate the desired CSDP synaptic adjustment. Our current results indicate that our design for systems of leaky integrate-and-fire neurons, connected via memristor-based synapses that are dynamically adapted via CSDP, are capable of yielding positive performance on a simple logical operation task. Future work will need to investigate our proof-of-concept's application to more complex tasks as well as compare its capabilities to alternative biological credit assignment schemes, such as feedback alignment and spike-timing-dependent plasticity.

\bibliographystyle{IEEEtran}
\footnotesize
\bibliography{refs.bib}

\end{document}